\documentclass[10pt,twocolumn,letterpaper]{article}

\usepackage{cvpr}              

\definecolor{cvprblue}{rgb}{0.21,0.49,0.74}
\usepackage[pagebackref,breaklinks,colorlinks,allcolors=cvprblue]{hyperref}
\usepackage[linesnumbered,ruled,vlined]{algorithm2e}
\usepackage{multirow}
\usepackage{pifont}
\usepackage[accsupp]{axessibility}

\def\paperID{3628} 
\def\confName{CVPR}
\def\confYear{2025}

\title{Efficient Test-time Adaptive Object Detection via Sensitivity-Guided Pruning}

\author{Kunyu Wang, Xueyang Fu, Xin Lu, Chengjie Ge, Chengzhi Cao, Wei Zhai, Zheng-Jun Zha$^{\ast}$ \\
School of Information Science and Technology and \\
MoE Key Laboratory of Brain-inspired Intelligent Perception and Cognition, \\
University of Science and Technology of China, Hefei, 230026, China \\
{\tt\small $\{$kunyuwang, luxion, cjge, chengzhicao$\}$@mail.ustc.edu.cn}\\
{\tt\small $\{$xyfu, wzhai056, zhazj$\}$@ustc.edu.cn}
}

\begin{document}
\maketitle
\begin{abstract}
Continual test-time adaptive object detection (CTTA-OD) aims to online adapt a source pre-trained detector to ever-changing environments during inference under continuous domain shifts. Most existing CTTA-OD methods prioritize effectiveness while overlooking computational efficiency, which is crucial for resource-constrained scenarios. In this paper, we propose an efficient CTTA-OD method via pruning. Our motivation stems from the observation that not all learned source features are beneficial; certain domain-sensitive feature channels can adversely affect target domain performance. Inspired by this, we introduce a sensitivity-guided channel pruning strategy that quantifies each channel based on its sensitivity to domain discrepancies at both image and instance levels. We apply weighted sparsity regularization to selectively suppress and prune these sensitive channels, focusing adaptation efforts on invariant ones. Additionally, we introduce a stochastic channel reactivation mechanism to restore pruned channels, enabling recovery of potentially useful features and mitigating the risks of early pruning. Extensive experiments on three benchmarks show that our method achieves superior adaptation performance while reducing computational overhead by 12$\%$ in FLOPs compared to the recent SOTA method.
\end{abstract}

\renewcommand{\thefootnote}{}
\footnotetext{$^{\ast}$Corresponding author. This work was supported by the National Natural Science Foundation of China (NSFC) under Grants 62225207, 62436008, 62422609 and 62276243.}    
\section{Introduction}
\begin{figure}[t]
    \centering
    \includegraphics[width=\linewidth]{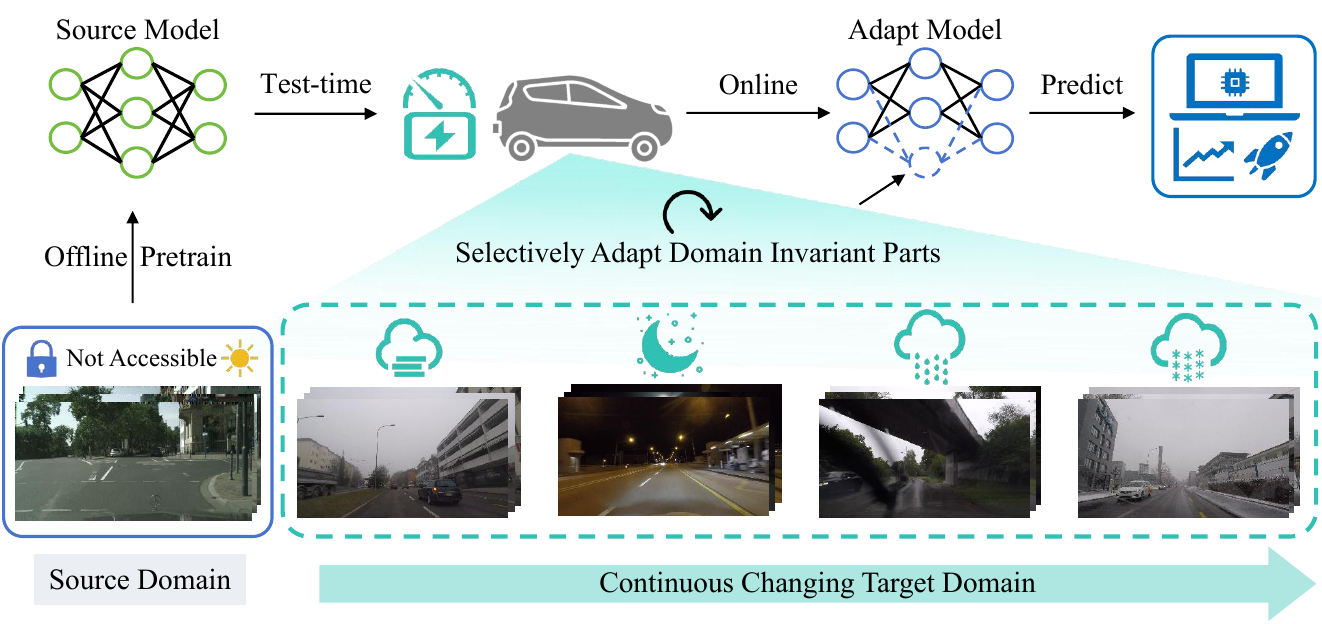}
    \caption{We consider continual test-time adaptive object detection, which adapts a source detector to changing environments during inference. By focusing adaptation efforts on invariant parts while pruning sensitive parts, we improve computational efficiency, which is crucial for resource-constrained scenarios like autonomous driving or UAV.}
    \label{intro_task}
    \vspace{-0.6cm}
\end{figure}

Object detection \cite{wang2023generalized,wang2024towards} is a fundamental task in computer vision with numerous downstream applications. However, in real-world scenarios such as autonomous driving \cite{uccar2017object} or robotics \cite{zhang2024navid,zhang2024uni}, various natural factors, such as adverse weather conditions and lighting changes \cite{ge2024neuromorphic,ge2022learning}, can lead to substantial domain shifts between training and testing data, resulting in significant performance degradation. Moreover, due to the dynamic nature of real-world environments, the test data distribution is constantly changing and inherently unpredictable, posing additional challenges. To address this issue, continual test-time adaptation \cite{wang2022continual} has been proposed, which aims to adapt a source pre-trained model during inference to accommodate the evolving test data, providing a promising solution to mitigate the domain shift problem, as shown in Fig \ref{intro_task}.

While existing continual test-time adaptive object detection methods show impressive results \cite{sinha2023test,chen2023stfar,mirza2023actmad}, enhancing computational efficiency has been relatively under-explored, despite its critical importance for resource-constrained scenarios \cite{liu2019edge,abrar2021energy,telli2023comprehensive}.
Current methods typically adapt all features learned from the source domain indiscriminately during adaptation to the target domain. However, our exploratory experiments reveal that not all learned source features are beneficial for target domain; in fact, certain source feature channels negatively affect the model's performance in the target domain.

\begin{figure}[t]
    \centering
    \includegraphics[width=0.95\linewidth]{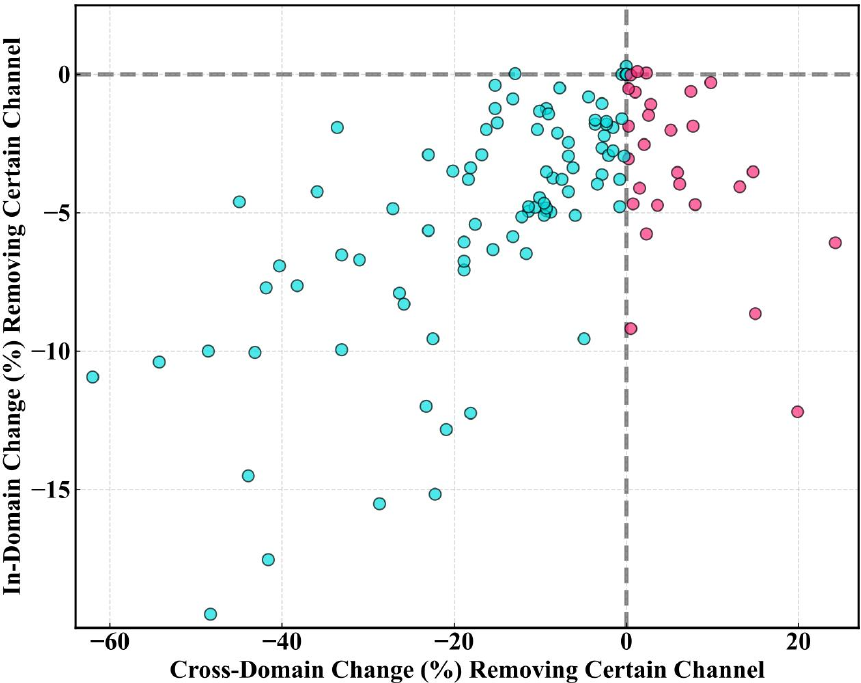}
    \caption{We analyze how removing certain feature channel from the source model affects in-domain and cross-domain performance. The source model is trained on the Cityscapes training set and evaluated in-domain on the Cityscapes validation set and cross-domain on Cityscapes-C. Points on the positive x-axis (or y-axis) indicate that removing certain channel improves cross-domain (or in-domain) performance, while those on the negative x-axis (or y-axis) indicate a decrease in cross-domain (or in-domain) performance when removed. The axis units denote the performance change percentage relative to direct-test.}
    \label{motivation}
    \vspace{-0.4cm}
\end{figure}

In particular, we perform ablation on feature channels of the source pre-trained model, observing the impact of removing certain channel on both in-domain and cross-domain performance. 
The results indicate that some feature channels in source model, marked as red points in Fig. \ref{motivation}, contribute positively to in-domain testing, where their removal causes a drop in in-domain performance. Conversely, these channels negatively impact cross-domain testing, as removing them enhances cross-domain performance. This observation reveals the an indiscriminate adaptation strategy is inefficient, as allocating computational resources to such domain-sensitive channels not only increases adaptation difficulty but also reduces efficiency.

Inspired by the success of network pruning in removing ineffective subsets to improve efficiency, we explore the potential of pruning similar ``ineffective subsets", namely feature channels sensitive to domain shifts, while focusing adaptation efforts on invariant ones to enhance adaptation efficiency.
To this end, we introduce a sensitivity-guided channel pruning that quantifies the weight of each feature channel based on its sensitivity to domain discrepancies. Specifically, we compute channel sensitivity weights at both the image level and instance level by measuring the discrepancy between target domain features and pre-collected source domain statistics. Subsequently, we apply a weighted sparsity regularization to the learnable parameters, encouraging the pruning of channels with higher sensitivity weights according to predefined threshold and ratio. Moreover, to prevent the loss of potentially useful channels due to early pruning, we design a stochastic channel reactivation mechanism. This mechanism utilizes Bernoulli sampling to randomly restore pruned channels, allowing the model to reassess their utility. By jointly optimizing the adaptation loss and the channel pruning loss, our method focuses on invariant features, thereby ensuring superior adaptation performance on the changing target domain while significantly reducing computational overhead.

\begin{figure}[t]
    \centering
    \includegraphics[width=0.95\linewidth]{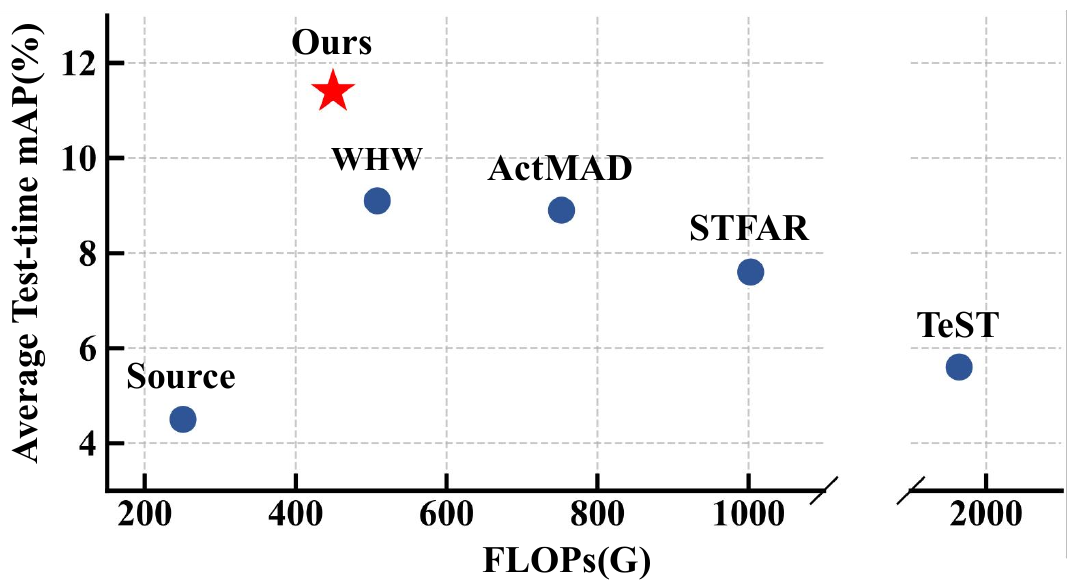}
    \caption{We perform continual online adaptation on Cityscapes $\rightarrow$ Cityscapes-C. The x-axis and y-axis denote the average test-time mAP across all corruptions and the total FLOPs consumption (including both forward and backward passes). Our method achieves superior results with the least FLOPs consumption.}
    \label{intro_result}
    \vspace{-0.5cm}
\end{figure}

In summary, our contributions are listed as follows:
\begin{itemize}
\item We unveil a novel perspective for enhancing CTTA-OD efficiency from the model structure viewpoint, inspired by the observation that certain source sensitive feature channels negatively impact target domain performance.

\item We propose a novel efficient CTTA-OD framework that suppresses and prunes sensitive feature channels and selectively adapts invariant ones, reducing adaptation difficulty and enhancing computational efficiency.

\item We propose a sensitivity-guided channel pruning strategy that quantifies each channel’s sensitivity to domain discrepancies at both image and instance levels, along with a stochastic reactivation mechanism to prevent early pruning of useful channels.
\end{itemize}

Extensive experiments on three benchmarks show that our method achieves superior adaptation results while reducing 12$\%$ FLOPs compared to the SOTA method, as shown in Fig \ref{intro_result}.

\section{Related Work}
\label{sec:related_work}

\begin{figure*}[t]
    \centering
    \includegraphics[width=\linewidth]{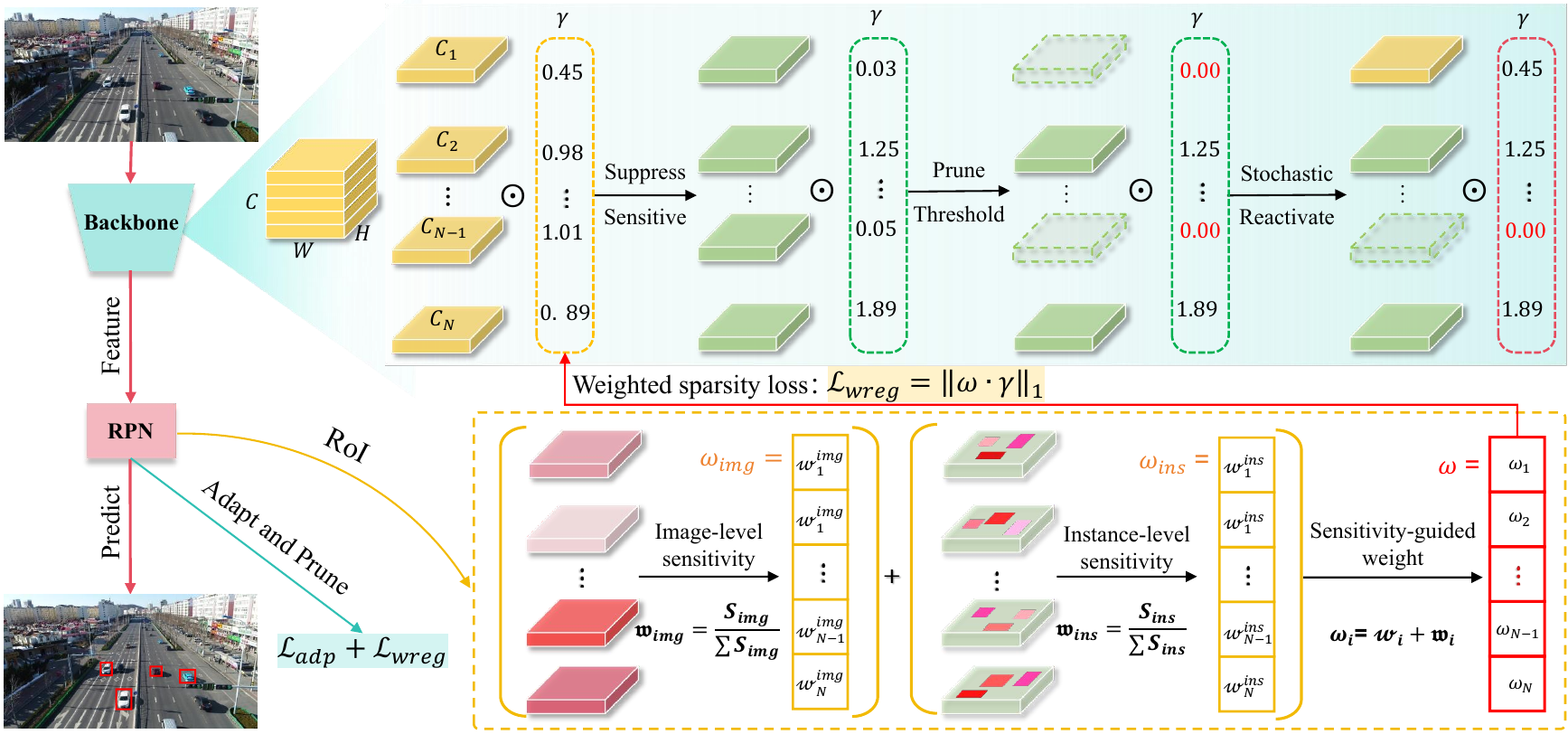}
    \caption{Overview of our method. Motivated by the observation that certain feature channels learned from the source domain negatively impact cross-domain performance, we propose suppressing and pruning sensitive channels while adapting invariant ones. We introduce sensitivity-guided channel pruning, which quantifies the importance of each feature channel based on its image-level and instance-level sensitivity to domain discrepancies. We then apply a weighted sparsity loss to the learnable parameters, promoting the pruning of channels with higher sensitivity based on the predefined threshold and ratio. To prevent the loss of potentially useful channels from early pruning, we introduce a stochastic channel reactivation mechanism.}
    \label{main}
    \vspace{-0.5cm}
\end{figure*}

\noindent
{\bf Test-time Adaptive Object Detection} (TTA-OD) \cite{deng2024balanced,khanh2024dynamic,yoon2024enhancing,zhao2024multi,chen2023exploiting,li2021free,chu2023adversarial,vs2023instance,liu2023periodically}
aims to adapt a source pre-trained detector to an unlabeled target domain without access to the original source data, addressing privacy concerns. To mitigate the absence of source data, most TTA-OD approaches adopt a self-training paradigm. For instance, 
SED \cite{li2021free} is the first approach to introduce pseudo-label self-training via self-entropy descent.
${A^2}$SFOD \cite{chu2023adversarial} incorporates an adversarial module into a mean teacher framework to align feature spaces between source-similar and source-dissimilar images. 
TTA-OD primarily addresses adaptation to the static target domain in an offline setting, assuming that test data follows an i.i.d. distribution with full access to the entire test set. However, in practice, test data is often streamed from continuously changing environments. Thus, we focus on the online continual setting, which is more aligned with real-world scenarios.

\smallskip
\noindent
{\bf Continual Test-time Adaptive Object Detection} (CTTA-OD) \cite{sinha2023test,mirza2022norm,chen2023stfar,mirza2023actmad,ruan2024fully,cao2024exploring,yoo2024and}
has gained increasing attention due to its practicality. DUA \cite{mirza2022norm} continuously adapts batch normalization statistics to refine feature representations. STFAR \cite{chen2023stfar} follows a self-training paradigm, generating pseudo-labeled objects and incorporating feature alignment regularization to improve the robustness of self-training. ActMAD \cite{mirza2023actmad} focuses on fine-grained alignment of activation statistics between test and training data. 
Most CTTA-OD methods primarily focus on improving effectiveness, while rarely exploring ways to enhance computational efficiency, despite the task's online requirement. 
In this paper, we offer a new perspective on enhancing computational efficiency from the model's viewpoint. By suppressing and pruning sensitive feature channels while adapting invariant ones, we achieve both effectiveness and efficiency.
\smallskip

\noindent
{\bf Network Pruning} \cite{cheng2024survey,lin2017runtime,hua2019channel,he2018soft,you2019gate,li2021dynamic,gao2018dynamic,liu2017learning}
aims to remove relatively ineffective subsets of parameters or structures from a neural network, improving efficiency while minimizing the degradation in model performance. Pruning methods are broadly categorized into unstructured and structured types. Unstructured pruning removes individual weights, creating sparsity but often requiring specialized hardware for acceleration. Structured pruning \cite{he2023structured}, by removing entire filters or layers, offers practical acceleration without such dependencies. Channel pruning, a key structured method, boosts computational efficiency by selectively removing feature channels. 
Motivated by the success of network pruning in reducing computational costs, we are prompted to explore whether similar ``ineffective subsets” exist in source pre-trained model. Our exploratory experiments reveal that certain sensitive feature channels in the source pre-trained network negatively impact target domain performance. This finding underscores the feasibility and potential of incorporating pruning to reduce computational overhead.
\section{Methodology}

In this section, we begin by introducing the task setup. We then describe our sensitivity-guided channel suppression and pruning. Following this, we present the optimization objectives and the pruning process, analyzing their effects on model computational efficiency. Finally, we introduce the stochastic channel reactivation strategy, which enables the model to reassess the utility of pruned channels. Fig \ref{main} provides an overview of our method.

\subsection{Problem Definition}
Assume we have an object detector pre-trained on the source domain $D_{s} = \{x_s, y_s\}$, where $x_s$ denotes the source images and $y_s$ denotes the corresponding bounding boxes and category labels. 
Our goal is to adapt the detector to a sequence of continuously changing target domains $\{D_{t}^1, D_{t}^2, ..., D_{t}^N\}$ using only the target data while making predictions. The target domain at time period $n$ is denoted as $D_{t}^{n} = \{x_t^{n}\}$, where $x_t^{n}$ denotes the target images at time period $n$ and $P_{test}^n \neq P_{test}^{n-1}$. 
Following prior work \cite{chen2023stfar,mirza2023actmad,yoo2024and}, as the source domain is inaccessible during adaptation, pre-computed source feature statistics, such as the mean and variance, are available.

\subsection{Network Pruning}
To achieve channel-level pruning, we utilize the learnable scaling factors in batch normalization (BN) layers to effectively identify and prune sensitive channels in the network, following prior work \cite{liu2017learning}. This approach offers two main advantages:
First, the ResNet backbone is composed of convolutional layers, BN layers, and activation functions. The BN layers normalize the features extracted by the convolutional layers and apply learnable scaling and shifting parameters to each channel:
\begin{equation}
    \hat{z} = \frac{z_{in} - \mu_B}{\sqrt{\sigma_B^2 + \epsilon}}, \quad z_{out} = \gamma \hat{z} + \beta,
\end{equation}
where $z_{in}$ and $z_{out}$ denote the input convolutional feature and output, respectively. $\mu_{\mathcal{B}}$ and $\sigma_{\mathcal{B}}$ are the mean and standard deviation values over $\mathcal{B}$.
The parameters $\gamma$ and $\beta$, which are trainable affine transformation factors, allow for the linear transformation of the normalized activations back to any scale.
By adjusting the scaling factors $\gamma$ in the BN layers, we can directly control the magnitude of each feature channel, enabling effective suppression and pruning of specific channels by reducing their scaling factors.

Furthermore, we apply $\mathcal{L}_1$ regularization as the sparsity loss $\mathcal{L}_{\text{reg}}$, pushing all scaling factors $\gamma_i$ to approach zero for channel suppression:
\begin{equation}
    \mathcal{L}_{\text{reg}} = \sum_{i=1}^{n} \|\gamma_i\|_1,
\end{equation}
where n denotes the total number of BN layers considered.
\subsection{Sensitivity-guided Channel Pruning}
Evidently, the above sparsity regularization loss suppresses all channels uniformly. 
However, based on our earlier analysis, we need to suppress domain-sensitive channels while preserving and adapting domain-invariant ones. 
To achieve this, we introduce a sensitivity-guided weighted sparsity loss $\mathcal{L}_{\text{wreg}}$, where the $\omega_i$ reflects the domain sensitivity of each channel $\gamma_i$:
\begin{equation}
\label{eq:loss_3}
    \mathcal{L}_{\text{wreg}} = \sum_{i=1}^{n} \|\omega_i \cdot \gamma_i\|_1,
\end{equation}
where $\cdot$ denotes element-wise multiplication.
Specifically, omitting the subscript $i$ for simplicity, $\omega$ consists of two components: image-level channel sensitivity $\omega_{\text{img}}$ and instance-level channel sensitivity $\omega_{\text{ins}}$:
\begin{equation}
    \omega = \omega_{\text{img}} + \omega_{\text{ins}}.
\end{equation}
We compute $\omega_{\text{img}}$ to capture image-level channel variation under domain shift as follows:
\begin{equation}
    S_{\text{img}} = \frac{\sum_{n=1}^{N} \left\| F_t^n - \overline{F_s} \right\|_1}{N D} \in \mathbb{R}^C,
\end{equation}
\begin{equation}
    w_{\text{img}} = C \times \frac{S_{\text{img}}}{\sum S_{\text{img}}},
\end{equation}
where $F_t^n$ denotes the feature map extracted by the convolutional layer before the specified BN layer for the $\text{n}$-th image in the target domain mini-batch, and $\overline{F_s}$ denotes the pre-computed average feature map at the same layer position from the source domain. Both $F_t^n$ and $\overline{F_s} \in \mathbb{R}^{\text{C} \times \text{D}}$, where N is the batch size, C is the number of channels, and D is the product of the H and W of the feature map. To stabilize training, we further normalize $S_{\text{img}}$ to balance the magnitude of variance across all channels, yielding $w_{\text{img}}$.

To capture fine-grained channel variation, we compute the instance-level $\omega_{\text{ins}}$ by leveraging the Regions of Interest (RoIs) predicted by the object detector on the target domain data. Specifically, we focus on RoIs with a background confidence less than 0.5, indicating potential foreground objects.
For each selected RoI in the $n$-th image of the target domain mini-batch, we extract the corresponding feature map region from the convolutional feature map $F_t^n$. 
For all obtained RoI features $\{f_{t}^{_m}\}_{m=1}^M$,  we utilize
the RoI-Alignment operation \cite{he2017mask} to align the spatial size of all instance-level features. We then calculate $\omega_{\text{ins}}$ as follows:
\begin{equation}
    S_{\text{ins}} = \frac{\sum_{m=1}^{M} \left\| f_{t}^{m} - \overline{f_s} \right\|_1}{M D_{\text{RoI}}} \in \mathbb{R}^C,
\end{equation}
\begin{equation}
    w_{\text{ins}} = C \times \frac{S_{\text{ins}}}{\sum S_{\text{ins}}},
\end{equation}
where $M$ denote the number of selected RoIs within the target domain mini-batch, and $D_{\text{RoI}}$ represent the spatial dimensions of the RoI features after alignment, $\overline{f_s} \in \mathbb{R}^{\text{C} \times \text{D}_{\text{RoI}}}$ refers to the pre-computed average RoI feature at the same layer in the source domain, based on predicted RoIs and their confidence scores. To ensure numerical stability, we also apply channel-wise normalization, yielding $w_{\text{ins}}$.

\begin{algorithm}[t]
\caption{Efficient CTTA-OD via pruning}
\label{alg:pruning}
\KwIn{Target data $x_t$, pruning threshold $t$, pruning ratio threshold $p$, reactivation probability $r$}
\For{each batch of target data $x_t^n$}{
    Prune channels where the BN scaling factor $\gamma < t$\ and define the computation graph\;
    Forward propagate for prediction\;
    Compute pruning ratio $\rho$ according to Eq. \eqref{eq:pruning_ratio}\;
    \eIf{$\rho < p$}{
        Compute loss as in Eq. \eqref{eq:loss_3} and Eq. \eqref{eq:loss_11}\;
    }{
        Compute loss as in Eq. \eqref{eq:loss_11}\;
        Stochastic pruned channels reactivation with $r$ according to Eq. \eqref{eq:reactivation}\;
    }
    Backward propagate for updating\;
}
\end{algorithm}

\begin{table*}[htbp]
\setlength{\tabcolsep}{2pt}
\renewcommand{\arraystretch}{1.1}
\centering
\caption{Comparisons on Cityscapes $\rightarrow$ Cityscapes-C. We report continual test-time adaptive detection results along with FLOPs consumption. 'Avg' denotes the average results over all conditions across 10 rounds. 'Fwd' indicates FLOPs during forward propagation, 'Bwd' indicates FLOPs during backward propagation, and 'Total' denotes the overall FLOPs.}
\label{tab:cityc}
\resizebox{\linewidth}{!}{
    \begin{tabular}{lcccccccccccccc@{\hskip 5pt}c@{\hskip 5pt}c}
    \toprule
    Round & \multicolumn{4}{c}{1} & \multicolumn{4}{c}{5} & \multicolumn{4}{c}{10} & \multirow{2}{*}{Avg} & \multicolumn{3}{c}{FLOPs} \\
    \cmidrule(r){1-13}
    \cmidrule(l){15-17}
    Condition & Motion & Snow & Defocus & Contrast & Motion & Snow & Defocus & Contrast & Motion & Snow & Defocus & Contrast & & Fwd & Bwd & Total \\
    \midrule
    Direct Test & 5.6 & 0.6 & 10.8 & 1.0 & 5.6 & 0.6 & 10.8 & 1.0 & 5.6 & 0.6 & 10.8 & 1.0 & 4.5 & 250.7 & 0.0 & 250.7 \\
    TeST \cite{sinha2023test} & 6.7 & 1.9 & 11.8 & 2.2 & 6.9 & 1.1 & 12.1 & 2.1 & 6.2 & 1.6 & 11.6 & 2.4 & 5.6 & 1011.1 & 953.2 & 1964.2 \\
    DUA \cite{mirza2022norm} & 6.0 & 0.8 & 11.3 & 1.4 & 6.1 & 0.9 & 11.1 & 1.2 & 5.2 & 0.2 & 10.5 & 0.5 & 4.6 & 250.7 & 0.0 & 250.7 \\
    STFAR \cite{chen2023stfar} & 7.5 & 1.9 & 12.4 & 2.4 & 10.2 & 2.7 & 12.4 & 5.0 & 10.6 & 2.7 & 11.7 & 4.1 & 7.6 & 501.4 & 501.4 & 1002.9 \\
    ActMAD \cite{mirza2023actmad} & 12.4 & 1.8 & 15.1 & 5.5 & 14.0 & 3.3 & 13.9 & 7.7 & 12.1 & 2.2 & 13.7 & 5.0 & 8.9 & 250.7 & 501.4 & 752.2 \\
    WHW \cite{yoo2024and} & 13.3 & 2.8 & 16.5 & 6.7 & 14.7 & 3.5 & 14.7 & 9.9 & 13.6 & 3.1 & 13.7 & 6.6 & 9.1 & 253.0 & 255.2 & 508.2 \\
    \midrule
    Ours & 13.2 & 2.6 & 17.1 & 7.4 & 15.6 & 4.6 & 15.1 & 10.3 & 15.2 & 4.8 & 15.1 & 10.4 & 11.4 & 224.6 & 225.0 & 449.6 \\
    \bottomrule
    \end{tabular}
}
\vspace{-0.4cm}
\end{table*}

\subsection{Optimization}
For the adaptation loss, we follow prior work \cite{chen2023stfar,mirza2023actmad,yoo2024and} and employ feature distribution alignment.
Specifically, we use both image-level and instance-level distribution alignment to minimize the KL divergence between source and target feature distributions. Let $\mu_{\text{s}}$ and $\Sigma_{\text{s}}$ denote the mean and variance of the source features. To capture distribution shifts in the test domain, we estimate the mean of the test features, $\mu_{\text{t}}$, using an exponentially moving average (EMA). The image-level alignment loss is defined as the KL divergence between the source and target distributions:
\begin{equation} 
    \mathcal{L}_{\text{img}} = D_{\text{KL}}\left(\mathcal{N}(\mu_{\text{s}}, \Sigma_{\text{s}}), \mathcal{N}(\mu_{\text{t}}, \Sigma_{\text{s}}) \right). 
\end{equation}
For instance-level alignment, we introduce an intra-class feature alignment loss that minimizes the KL divergence between the source and target distributions for each category:
\begin{equation} 
    \mathcal{L}_{\text{ins}} = \sum_{k} w_{k} \cdot D_{\text{KL}}\left(\mathcal{N}(\mu^{k}_{\text{s}}, \Sigma^{k}_{\text{s}}), \mathcal{N}(\mu^{k}{\text{t}}, \Sigma^{k}_{\text{s}}) \right),
\end{equation}
where $\mu^{k}_{\text{s}}$ and $\Sigma^{k}_{\text{s}}$ are the mean and variance for category $k$ in the source domain, and $\mu^{k}_{\text{t}}$ denotes the EMA-updated mean for category $k$ in the target domain.
To further address class imbalance, we dynamically adjust the weight $w_{k}$ for each category based on its frequency in the target domain, assigning higher weights to rare classes to enhance alignment quality. Overall, the adaptation loss is as follows:
\begin{equation} 
\label{eq:loss_11}
    \mathcal{L}_{\text{adp}} = \mathcal{L}_{\text{img}} + \mathcal{L}_{\text{ins}}. 
\end{equation}

During adapting the continuously incoming target data, our method dynamically prunes channels for all the considered BN layers before each optimization step, constructing a sub-network for the current step. Specifically, with a predefined pruning threshold $t$, for each scaling parameter $\gamma_i$ in the BN layers, if $\gamma_i < t$, the corresponding channel and the associated filter in the preceding convolutional layer are removed, defining the forward computation graph for the current step.
We define the pruning ratio $\rho$ as the ratio of pruned channels (those satisfying $\gamma_i < t$) to the total number of channels across considered BN layers:
\begin{equation}
\label{eq:pruning_ratio}
    \rho = \frac{\sum_i |\{\gamma_i \mid \gamma_i < t\}|}{\sum_i |\{\gamma_i\}|}.
\end{equation}
During the backward optimization process, the loss function $\mathcal{L}_{total}$ is defined based on the pruning ratio $\rho$ as follows:
\begin{equation}
    \mathcal{L}_{total} = \begin{cases} 
      \mathcal{L}_{\text{adp}} + \lambda \mathcal{L}_{\text{wreg}}, & \text{if } \rho < p \\
      \mathcal{L}_{\text{adp}}, & \text{if } \rho \ge p 
   \end{cases}
\end{equation}
where $\lambda$ is a balancing coefficient between the adaptation loss and the pruning loss, $p$ is the predefined pruning ratio threshold.
By jointly optimizing with the adaptation loss and regulating the pruning ratio through the threshold $p$, we effectively prevent over-pruning, balancing adaptability and computational efficiency.

\noindent
{\bf Discussion about efficiency.}
The ResNet backbone comprises stacked modules like BasicBlock and Bottleneck, each with multiple Conv-BN-ReLU groupings. Except for the down-sampling group in the residual branch and the first group in the main branch, other convolution layers exhibit both input and output sparsity. Positioned between two BN layers, these layers support efficient pruning by removing channels within each filter as well as eliminating entire filters. This setup results in a quadratic gain in computational efficiency relative to the pruning ratio; for instance, retaining only half of the channels in both adjacent BN layers reduces the computational load of the intermediate convolution layer to one-fourth of its original cost.

Our method reduces computational costs in both forward and backward propagation. In forward propagation, fewer nodes lead to reduced computational requirements. Backpropagation costs primarily arise from gradient propagation through the computation graph and the calculation of gradients for updatable parameters. By shortening gradient paths and reducing node count, the simplified computation graph lowers the computational burden of gradient propagation. Furthermore, with fewer parameters involved, the cost of computing their gradients is also minimized.

\subsection{Stochastic Channel Reactivation}
During optimization, some channels may be subject to early pruning or mis-pruning due to initial scaling parameter values or the dynamic nature of the target domain. Once a channel is pruned, it is excluded from subsequent forward and backward passes, eliminating any chance to correct pruned channels. To this end, we propose a stochastic channel reactivation strategy.
Specifically, when the pruning ratio $\rho$ exceeds the predefined pruning ratio threshold $p$, we employ Bernoulli sampling to determine whether each pruned channel should be reactivated. For each pruned channel $c_i^j$, a random variable $b_i^j$ is sampled from a Bernoulli distribution:
\begin{equation}
\label{eq:reactivation}
    b_i^j \sim \text{Bernoulli}(r),
\end{equation}
where $r$ is a small reactivation probability. If $b_i^j=1$, the channel's BN scaling parameter $\gamma_i^j$ is reset to its initial pre-trained value from the source domain. 
This reactivation allows the channel to rejoin the model's forward and backward passes, giving the model an opportunity to reassess the utility of the reactivated channels, mitigating losses caused by early pruning or mis-pruning.
For clarity, the workflow of our method is illustrated in Algorithm \ref{alg:pruning}.
\section{Experiments}

\begin{table*}[htbp]
\setlength{\tabcolsep}{3pt}
\renewcommand{\arraystretch}{1.1}
\centering
\caption{Comparisons on UAVDT $\rightarrow$ UAVDT-C. We report continual test-time adaptive detection results along with FLOPs consumption.}
\label{tab:uavdt}
\resizebox{\linewidth}{!}{
    \begin{tabular}{lcccccccccccccc@{\hskip 5pt}c@{\hskip 5pt}c}
    \toprule
    Round & \multicolumn{4}{c}{1} & \multicolumn{4}{c}{5} & \multicolumn{4}{c}{10} & \multirow{2}{*}{Avg} & \multicolumn{3}{c}{FLOPs} \\
    \cmidrule(r){1-13}
    \cmidrule(l){15-17}
    Condition & Fog & Defocus & Motion & Night & Fog & Defocus & Motion & Night & Fog & Defocus & Motion & Night & & Fwd & Bwd & Total \\
    \midrule
    Direct Test & 4.8   & 2.5   & 6.8   & 11.4  & 4.8   & 2.5   & 6.8   & 11.4  & 4.8   & 2.5   & 6.8   & 11.4  & 6.4   & 250.7  & 0.0   & 250.7  \\
    TeST \cite{sinha2023test}  & 6.4   & 4.5   & 10.5  & 12.0  & 8.2   & 6.9   & 10.2  & 12.1  & 8.8   & 6.6   & 10.5  & 12.2  & 9.2   & 1011.1  & 953.2  & 1964.2  \\
    DUA \cite{mirza2022norm}  & 5.2   & 2.8   & 7.1   & 11.7  & 5.0   & 2.8   & 6.9   & 11.8  & 3.6   & 1.1   & 5.5   & 10.3  & 6.1   & 250.7  & 0.0   & 250.7  \\
    STFAR \cite{chen2023stfar} & 6.9   & 5.7   & 11.3  & 12.3  & 9.5   & 6.9   & 10.5  & 12.0  & 9.0   & 6.6   & 10.8  & 12.2  & 9.6   & 501.4  & 501.4  & 1002.9  \\
    ActMAD \cite{mirza2023actmad}& 9.5   & 6.4   & 15.3  & 15.0  & 13.2  & 9.1   & 13.4  & 14.0  & 13.1  & 7.5   & 12.3  & 13.9  & 10.8  & 250.7  & 501.4  & 752.2  \\
    WHW \cite{yoo2024and}  & 9.9   & 7.8   & 17.0  & 15.6  & 16.1  & 10.1  & 14.6  & 14.9  & 13.9  & 7.1   & 13.6  & 13.4  & 12.0  & 253.0  & 255.2  & 508.2  \\
    \midrule
    Ours  & 10.2  & 8.9   & 17.2  & 16.9  & 16.6  & 11.1  & 16.0  & 15.4  & 18.0  & 10.1  & 16.3  & 15.5  & 14.3  & 221.0  & 221.4  & 442.4  \\
    \bottomrule
    \end{tabular}%
}
\vspace{-0.1cm}
\end{table*}

\begin{table*}[htbp]
\setlength{\tabcolsep}{5pt}
\renewcommand{\arraystretch}{1.1}
\centering
\caption{Comparisons on Cityscapes $\rightarrow$ ACDC. We report continual test-time adaptive detection results along with FLOPs consumption.}
\label{tab:acdc}
\resizebox{\linewidth}{!}{
    \begin{tabular}{lcccccccccccccc@{\hskip 5pt}c@{\hskip 5pt}c}
    \toprule
    Round & \multicolumn{4}{c}{1} & \multicolumn{4}{c}{5} & \multicolumn{4}{c}{10} & \multirow{2}{*}{Avg} & \multicolumn{3}{c}{FLOPs} \\
    \cmidrule(r){1-13}
    \cmidrule(l){15-17}
    Condition & Snow & Rain & Night & Fog & Snow & Rain & Night & Fog & Snow & Rain & Night & Fog & & Fwd & Bwd & Total \\
    \midrule
Direct Test & 21.8  & 22.2  & 9.8   & 34.4  & 21.8  & 22.2  & 9.8   & 34.4  & 21.8  & 22.2  & 9.8   & 34.4  & 22.0  & 250.7  & 0     & 250.7  \\
    TeST  \cite{sinha2023test}  & 22.0  & 22.1  & 9.6   & 34.4  & 21.8  & 22.3  & 9.6   & 34.1  & 22.0  & 22.2  & 9.5   & 34.4  & 22.1  & 1011.1  & 953.2  & 1964.2  \\
    DUA \cite{mirza2022norm}  & 20.9  & 21.5  & 9.1   & 33.8  & 21.1  & 21.3  & 9.0   & 33.7  & 20.4  & 20.8  & 8.4   & 33.1  & 21.1  & 250.7  & 0     & 250.7  \\
    STFAR \cite{chen2023stfar} & 21.9  & 22.5  & 9.5   & 34.7  & 21.7  & 22.2  & 9.5   & 34.2  & 21.6  & 22.8  & 9.6   & 34.7  & 22.1  & 501.4  & 501.4  & 1002.9  \\
    ActMAD \cite{mirza2023actmad} & 22.8  & 22.6  & 10.7  & 33.5  & 23.4  & 22.7  & 11.5  & 34.5  & 22.2  & 22.9  & 10.9  & 34.9  & 23.3  & 250.7  & 501.4  & 752.2  \\
    WHW \cite{yoo2024and}  & 23.2  & 22.6  & 11.2  & 34.7  & 24.0  & 24.1  & 12.4  & 34.2  & 24.0  & 23.6  & 12.7  & 34.7  & 23.6  & 253.0  & 255.2  & 508.2  \\
    \midrule
    Ours  & 23.4  & 22.6  & 11.7  & 34.7  & 24.7  & 24.1  & 13.0  & 35.1  & 24.5  & 24.1  & 12.9  & 35.0  & 24.2  & 226.2  & 226.6  & 452.8  \\
    \bottomrule
    \end{tabular}%
}
\vspace{-0.4cm}
\end{table*}

\begin{table}[htbp]
  \centering
  \caption{Ablation analysis of the framework components, where ’SCR‘ represents Stochastic Channel Reactivation.}
  \resizebox{0.9\linewidth}{!}{
    \begin{tabular}{ccccccc}
    \toprule
    $\mathcal{L}_{\text{adp}}$   & $\mathcal{L}_{\text{reg}}$   & $\omega_{\text{img}}$   & $\omega_{\text{ins}}$   & SCR   & Avg   & $\Delta$FLOPs$\downarrow$ \\
    \midrule
          &       &       &       &       & 4.5   & 0.0\%  \\
    \ding{51}     &       &       &       &       & 10.9  & 0.0\%  \\
    \ding{51}     & \ding{51}     &       &       &       & 8.8   & 10.2\%  \\
    \ding{51}     & \ding{51}     & \ding{51}     &       &       & 10.5  & 10.5\%  \\
    \ding{51}     & \ding{51}     &      &   \ding{51}   &       &  9.9  & 10.9\%  \\
    \ding{51}     & \ding{51}     & \ding{51}     & \ding{51}     &       & 11.2  & 10.7\%  \\
    \ding{51}     & \ding{51}     & \ding{51}     & \ding{51}     & \ding{51}     & 11.4  & 10.4\%  \\
    \bottomrule
    \end{tabular}%
    }
  \label{tab:component}%
  \vspace{-0.3cm}
\end{table}%

\begin{figure*}[t]
    \centering
    \includegraphics[width=\linewidth]{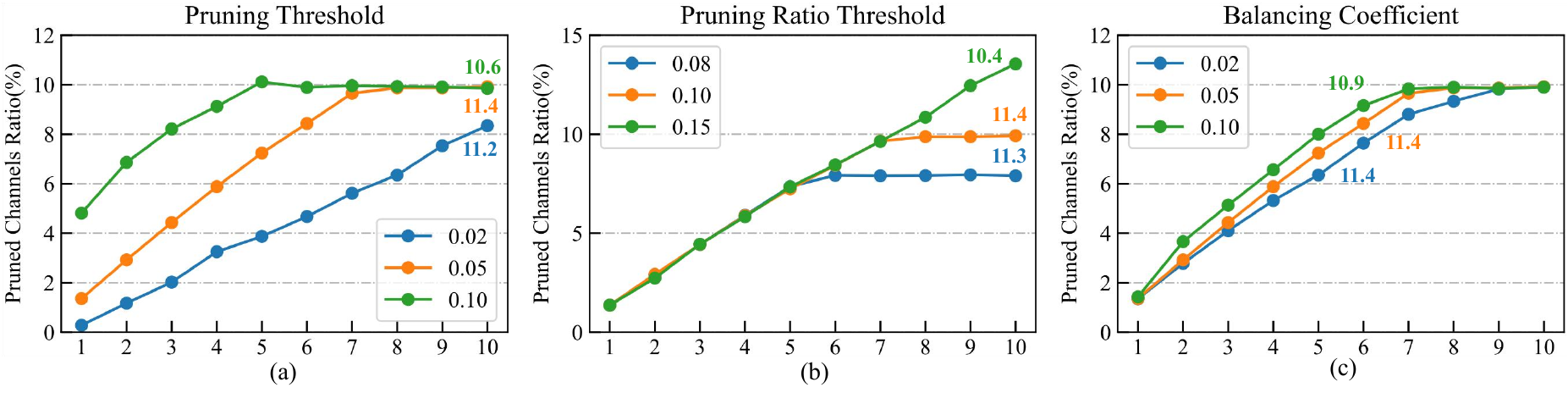}
    \vspace{-0.7cm}
    \caption{Ablation analysis of three hyper-parameters: (a) Pruning threshold $t$, (b) Pruning ratio threshold $p$, and (c) Balancing coefficient $\lambda$. We provide the pruned channel ratio for each round, with the average mAP for each setting annotated on the line plot.}
    \vspace{-0.2cm}
    \label{expr_hyper}
\end{figure*}

\begin{figure*}[t]
    \centering
    \includegraphics[width=\linewidth]{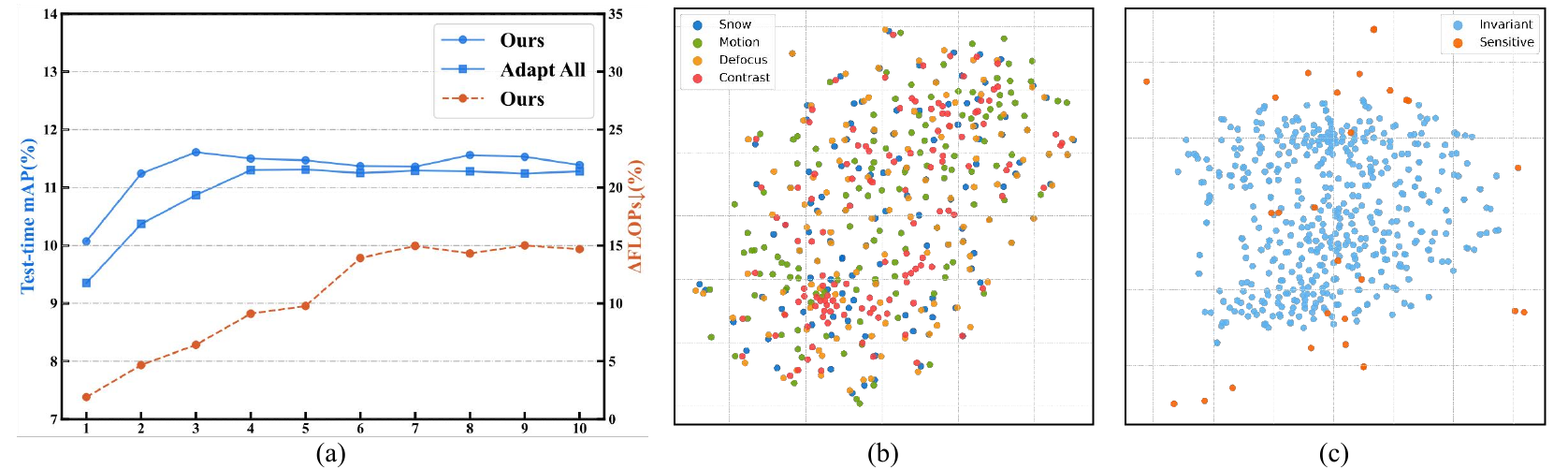}
    \vspace{-0.6cm}
    \caption{Visualization analysis of the effectiveness of pruning sensitive channels. (a) mAP change curves during test-time adaptation for our method versus the adapt-all-channels approach, with FLOPs reduction for our method. (b) T-SNE visualization of invariant channel distribution changes retained by our method in the final adaptation round across four target domains. (c) T-SNE visualization of feature channel distribution when adapting all channels in the final round, with sensitive and invariant channels color-coded based on our mask.}
    \label{expr_visualization}
    \vspace{-0.6cm}
\end{figure*}

\subsection{Experimental Setup}

\noindent
{\bf Datasets.}
We conduct experiments in both autonomous driving and UAV scenarios, including \textbf{Cityscapes-C}, \textbf{ACDC}, and \textbf{UAVDT-C}. 
The \textbf{Cityscapes} dataset \cite{cordts2016cityscapes} consists of 2,975 training images and 500 validation images with 8 categories of objects. We construct \textbf{Cityscapes-C} based on benchmark robustness tasks \cite{hendrycks2019benchmarking}, selecting four common corruption types: snow, contrast, motion blur, and defocus blur. These corruptions are applied to the validation set at the maximum severity level 5, with each corruption type forming an individual target domain consisting of 500 images.
The \textbf{ACDC} dataset \cite{sakaridis2021acdc} shares the same class categories as Cityscapes but includes four different adverse visual conditions: fog, night, rain, and snow, with each condition containing 400 unlabeled images.
The \textbf{UAVDT} dataset \cite{du2018unmanned} can be separated into three parts based on the weather annotations: daylight, nighttime, and fog. We use 17.5k images from the daylight part as the source domain and extract 500 images each from the nighttime and fog parts as target domains. Additionally, 500 images from the remaining daylight images are selected and applied with motion blur and defocus blur based on robustness benchmark \cite{hendrycks2019benchmarking}. Thus, the \textbf{UAVDT-C} include nighttime, fog, motion blur, and defocus blur.
For all experiment settings, we repeat 10 times adaptation to target domain group to evaluate performance.
\smallskip

\noindent
{\bf Implementation Details.} 
We use Faster R-CNN \cite{ren2016faster} with a ResNet18 \cite{he2016deep} backbone pre-trained on ImageNet \cite{deng2009imagenet} as the detector. During test-time adaptation, we adapt the learnable scaling factors in the BN layers while freezing all other parameters pre-trained on the source domain. The batch size is set to 4, and the learning rate for the Adam optimizer is set to 0.005. For hyper-parameter settings, the pruning threshold $t$ is set to 0.05, the pruning ratio threshold $p$ to 0.1, the reactivation probability $r$ to 0.01, and the loss balancing coefficient $\lambda$ to 0.05.
Following prior work \cite{wang2022continual}, we simulate the continual changing target domains via 10 rounds adaptation.

\smallskip
\noindent
{\bf Evaluation Metrics.} 
We use mAP@50 ($\%$) and FLOPs (G) as the evaluation metrics for detection performance and computational efficiency. 
FLOPs during adaptation mainly consist of two components: Forward propagation (Fwd) and Backward propagation (Bwd). Fwd FLOPs can be calculated based on the model structure or using auxiliary tools. In models where all parameters are learnable, Bwd typically requires twice the FLOPs of Fwd \cite{epoch2021backwardforwardFLOPratio}, mainly consisting of two parts: gradient propagation across layers and gradient computation for the learnable parameters.
Given that the Fwd of $f_{i+1} = f_i\mathcal{W} + b$, the Bwd from the $i+1^{th}$ layer to the $i^{th}$ layer, and the weight gradient are formulated as:
\begin{equation}
\label{eq:layer}
    \frac{\partial \mathcal{L}}{\partial f_i} = \frac{\partial \mathcal{L}}{\partial f_{i+1}} \mathcal{W}^T, 
\end{equation}
\begin{equation}
\label{eq:weight}
    \frac{\partial \mathcal{L}}{\partial \mathcal{W}} = f_i^T \frac{\partial \mathcal{L}}{\partial f_{i+1}},
\end{equation}
where Eq. \eqref{eq:layer} describes gradient propagation across layers, which must be computed regardless of whether the parameters in the current layer are learnable, and Eq. \eqref{eq:weight} describes the gradient computation for learnable parameters, which is only necessary if the parameters of the current layer are learnable.
Therefore, the FLOPs for Fwd can be calculated based on the model’s computation graph, and the FLOPs for Bwd can be calculated based on the FLOPs for Fwd and the distribution of learnable parameters, enabling calculation of the total FLOPs. FLOPs computation for convolutional layers is similar, so we skip this discussion.

\subsection{Benchmark Results}
We compare our approach with the baseline (directly testing the source model on target domains) and five competing CTTA-OD methods \cite{sinha2023test,mirza2022norm,chen2023stfar,mirza2023actmad,yoo2024and} across three benchmarks. 
For open-source methods, we strictly follow the provided hyper-parameters, and for others, we replicate based on paper details.
The results are shown in Tables \ref{tab:cityc}, \ref{tab:uavdt}, \ref{tab:acdc}. 
Our method achieves an increase of 2.3, 2.3, and 0.6 in ten-round average mAP over the recent SOTA method, while reducing FLOPs by 11.6\%, 12.9\%, and 10.9\%, respectively. Compared to the baseline, our method boosts average mAP by 6.9, 7.9, and 2.2 with the lowest additional FLOPs among all CTTA-OD methods.
In DUA, multi-round adaptation leads to severe disruption of BN layer statistics by domain shifts, which does not contribute positively to target domain performance. TeST and STFAR, based on the mean teacher framework, achieve stable improvements during adaptation but incur substantial additional computation due to the inference and updates required for a larger set of parameters. Compared to alignment-based methods such as ActMAD and WHW, our approach prunes sensitive channels, reducing adaptation difficulty. Additionally, the smaller number of optimized parameters allows for faster adaptation to changing target domains and minimizes error accumulation, resulting in the highest performance gains.

\subsection{Ablation Analysis}
\noindent
{\bf Framework Components.} 
We conduct an ablation study to verify the effectiveness of each component in our proposed framework, as detailed in Table \ref{tab:component}. Indiscriminate pruning leads to notable performance degradation, whereas channel sensitivity guidance at both image and instance levels enables the model to make informed pruning decisions, thereby preserving performance. Stochastic channel reactivation provides additional performance gain, albeit with a minor increase in the FLOPs consumption.
\smallskip

\noindent
{\bf Hyper-parameters.}
We investigate how varying the setting of pruning threshold $t$, pruning ratio threshold $p$, and balancing coefficient $\lambda$. affects the pruning process and adaptation performance, as shown in Fig \ref{expr_hyper}.
Setting a high pruning threshold risks early pruning, removing useful channels and causing performance loss, while a low threshold stabilizes performance but slows pruning process, limiting FLOPs reduction. A high pruning ratio threshold may lead to over-pruning, causing substantial performance degradation. A larger balancing coefficient between losses prioritizes pruning loss optimization, accelerating the pruning process; however, an aggressive pruning strategy may also compromise performance.

\subsection{Visualization Analysis}
To validate the effectiveness of pruning sensitive channels, we conduct visualization analysis. Fig \ref{expr_visualization} (a) compares mAP change curves during test-time adaptation between adapting all channels and our method, which focuses only on invariant channels. Pruning sensitive channels reduces adaptation difficulty, enabling faster learning of target domain distributions in early rounds and boosting average mAP. It also cuts computational overhead, while adapting all channels does not decrease FLOPs. Fig \ref{expr_visualization} (b) visualizes the distribution changes of the invariant channels retained by our method through T-SNE in the final round of continual adaptation across four target domains. The mixed distribution across these domains demonstrates the invariance of the feature channels. 
Fig \ref{expr_visualization} (c) presents the feature channel distribution when adapting all channels during the final round, with sensitive and invariant channels differentiated by color using our channel mask. Sensitive channels show scattered distributions under changing target domains, while invariant channels remain more stable, validating the necessity of pruning sensitive channels.
\section{Conclusion and Future Work}
\noindent
In this paper, we introduce an efficient CTTA-OD framework based on sensitivity-guided pruning, offering a new perspective on enhancing computational efficiency from a model structure standpoint. For future work, one promising direction is to develop a more subtle pruning strategy, as focusing solely on general invariant features may not fully exploit the model's potential. Furthermore, extending our approach to versatile detectors such as YOLO \cite{diwan2023object}, DETR \cite{carion2020end}, or models without BN layers will also be crucial for broader application..
{
    \small
    \bibliographystyle{ieeenat_fullname}
    \bibliography{main}
}

\end{document}